\documentclass[11pt,a4paper]{article}
\usepackage[hyperref]{emnlp2020}
\usepackage{times}
\usepackage{latexsym}
\usepackage{amsmath}
\usepackage{graphicx}
\usepackage{amssymb}
\usepackage{makecell}
\usepackage[noend]{algpseudocode}
\usepackage{algorithmicx,algorithm}
\def\aclpaperid{956}

\usepackage{microtype}

\aclfinalcopy 

\title{BERT-EMD: Many-to-Many Layer Mapping for BERT Compression with Earth Mover's Distance}

\author{Jianquan Li\textsuperscript{1}\thanks{\quad Equal contribution}, Xiaokang Liu\textsuperscript{1}\footnotemark[1], Honghong Zhao\textsuperscript{1}, Ruifeng Xu\textsuperscript{2}, Min Yang\textsuperscript{3}\thanks{\quad Min Yang is corresponding author},  Yaohong Jin\textsuperscript{1}\\ 
 \textsuperscript{1}Beijing Ultrapower Software Co.,Ltd., China \\ 
 \textsuperscript{2}Harbin Institute of Technology (Shenzhen), China \\ 
 \textsuperscript{3}Shenzhen Key Laboratory for High Performance Data Mining,\\Shenzhen Institutes of Advanced Technology, Chinese Academy of Sciences, China \\
\texttt{ \{lijianquan2,liuxiaokang1,zhaohonghong1,jinyaohong\}@ultrapower.com.cn}\\ \texttt{xuruifeng@hit.edu.cn}, \quad \texttt{min.yang@siat.ac.cn}
}

\begin{document}
\maketitle
\begin{abstract}
Pre-trained language models (e.g., BERT) have achieved significant success in various natural language processing (NLP) tasks. 
However, high storage and computational costs obstruct pre-trained language models to be effectively deployed on resource-constrained devices. In this paper, we propose a novel BERT distillation method based on many-to-many layer mapping, which allows each intermediate student layer to learn from any intermediate teacher layers.
In this way, our model can learn from different teacher layers adaptively for various NLP tasks.
In addition, we leverage Earth Mover's Distance (EMD) to compute the minimum cumulative cost that must be paid to transform knowledge from teacher network to student network. EMD enables the effective matching for many-to-many layer mapping. 
Furthermore, we propose a cost attention mechanism to learn the layer weights used in EMD automatically, which is supposed to further improve the model's performance and accelerate convergence time.
Extensive experiments on GLUE benchmark demonstrate that our model achieves competitive performance compared to strong competitors in terms of both accuracy and model compression. For reproducibility, we release the code and data at  \url{https://github.com/lxk00/BERT-EMD}.
\end{abstract}

\section{Introduction}
In recent years, pre-trained language models, such as GPT \cite{radford2018improving}, BERT \cite{devlin2018bert}, XL-Net \cite{yang2019xlnet}, have been proposed and applied to many NLP tasks, yielding state-of-the-art performances. 
However, the promising results of the pre-trained language models come with the high costs of computation and memory in inference, which obstruct these pre-trained language models to be deployed on  resource-constrained devices and real-time applications. For example, the original BERT-base model, which achieved great success in many NLP tasks, has 12 layers and about 110 millions parameters. 

It is therefore critical to effectively accelerate inference time and reduce the computational workload while maintaining accuracy. This research issue has attracted increasing attention \cite{wang2019structured,shen2019q,tang2019distilling}, of which knowledge distillation \cite{tang2019distilling} is considered to be able to provide a practical way. Typically, knowledge distillation techniques train a compact and shallow student network under the guidance of a complicated larger teacher network with a teacher-student strategy \cite{watanabe2017student}. Once trained, this compact student network can be directly deployed in real-life applications.

So far, there have been several studies, such as DistilBERT \cite{tang2019distilling}, BERT-PKD \cite{sun2019patient}, TinyBERT \cite{jiao2019tinybert}, which attempt to compress the original BERT into a lightweight student model without performance sacrifice based on knowledge distillation. For example, BERT-PKD \cite{sun2019patient} and  TinyBERT \cite{jiao2019tinybert} are two representative BERT compression approaches, which encourage the student model to extract knowledge from both the last layer and the intermediate layers of the teacher network. 

Despite the effectiveness of previous studies, there are still several challenges for distilling comprehensive knowledge from the teacher model, which are not addressed well in prior works.
\textbf{First}, existing compression methods learn one-to-one layer mapping, where each student layer is guided by only one specific teacher layer. For example, BERT-PKD uses the 2, 4, 6, 8, 10 teacher layers to guide the 1 to 5 student layers, respectively. However, these one-to-one layer mapping strategies are assigned based on empirical observations without theoretical guidance. 
\textbf{Second}, as revealed in \cite{clark2019does}, different BERT layers could learn different levels of linguistic knowledge.
The one-to-one layer mapping strategy cannot learn an optimal, unified compressed model for different NLP tasks. In addition, most previous works do not consider the importance of each teacher layer and use the same layer weights among various tasks, which create a substantial barrier for generalizing the compressed model to different NLP tasks. Therefore, an adaptive compression model should be designed to transfer knowledge from all teacher layers dynamically and effectively for different NLP tasks. 

To address the aforementioned issues, we propose a novel BERT compression approach based on many-to-many layer mapping and Earth Mover's Distance (EMD) \cite{rubner2000earth}, called BERT-EMD. \textbf{First}, we design a many-to-many layer mapping strategy, where each intermediate student layer has the chance to learn from all the intermediate teacher layers. In this way, BERT-EMD can learn from different intermediate teacher layers adaptively for different NLP tasks, motivated by the intuition that different NLP tasks require different levels of linguistic knowledge contained in the intermediate layers of BERT. 
\textbf{Second}, to learn an optimal many-to-many layer mapping strategy, we leverage EMD to compute the minimum cumulative cost that must be paid to transform knowledge from teacher network to student network. EMD is a well-studied optimization problem and provides a suitable solution to transfer knowledge from the teacher network in a holistic fashion.

We summarize our main contributions as follows. 
(1) We propose a novel many-to-many layer mapping strategy for compressing the intermediate layers of BERT in an adaptive and holistic fashion. 
(2) We leverage EMD to formulate the distance between the teacher and student networks, and learn an optimal many-to-many layer mapping based on a solution to the well-known transportation problem. 
(3) We propose a cost attention mechanism to learn the layer weights used in EMD automatically, which can further improve the model's performance and accelerate convergence time.
(4) Extensive experiments on GLUE tasks show that  BERT-EMD achieves better performance than the state-of-the-art BERT distillation methods.

\section{Related Work}
Language models pre-trained on large-scale corpora can learn universal language
representations, which have proven to be effective in many NLP tasks \cite{mikolov2013distributed, pennington2014glove, joulin2016bag}. 
Early efforts mainly focus on learning good word embeddings, such as word2vec \cite{mikolov2013distributed} and GloVe \cite{pennington2014glove}.  Although these pre-trained
embeddings can capture semantic meanings of words, they are context-free and fail to capture higher-level concepts in context, such as syntactic structures and polysemous disambiguation. 
Subsequently, researchers have shifted attention to contextual word embeddings learning, such as ELMo~\cite{peters2018deep}, ULMFit ~\cite{howard2018universal}, GPT~\cite{radford2018improving}, BERT~\cite{devlin2018bert}, ENRIE ~\cite{zhang2019ernie}, XL-Net~\cite{yang2019xlnet}, RoBERTa~\cite{liu2019roberta}.
For example, \citet{devlin2018bert} released the BERT-base of 110 million parameters and BERT-large of 330 million parameters, which achieved significantly better results than previous methods on GLUE tasks.

However, along with high-performance, the pre-trained language models (e.g., BERT) usually have a large number of parameters, which require a high cost of computation and memory in inference. Recently, many attempts have been made to reduce the computation overhead and model storage of pre-trained language models without performance sacrifice. Existing compression techniques can be divided into three categories: 
low-rank matrix factorization ~\cite{wang2019structured}, quantization ~\cite{shen2019q}, and knowledge distillation~\cite{tang2019distilling}. 
Next, we mainly review the related works that use knowledge distillation to compress the BERT model.

Knowledge distillation using the teacher-student strategy learns a lightweight student network under the guidance of a large and complicated teacher network. 
\citet{mukherjee2019distilling} distilled BERT into an LSTM network via both hard and soft distilling methods. 
\citet{sun2019patient} proposed the BERT-PKD model to transfer the knowledge from both the final layer and the intermediate layers of the teacher network.
\citet{jiao2019tinybert} proposed the TinyBERT model, which performed the Transformer distillation at both pre-training and fine-tuning processes.  
\citet{xu2020bert} proposed the BERT-of-Theseus model to learn a compact student network by replacing the teacher layers with their substitutes.
\citet{sun2020mobilebert} introduced the MobileBERT model, which has the same number of layers with the teacher network, but was much narrower via adopting bottleneck structures. 
\citet{wang2020minilm} distilled the self-attention module of the last Transformer layer of the teacher network. 

However, the aforementioned BERT compression approaches struggle to find an optimal layer mapping between the teacher and student networks. Each student layer merely learns from a single teacher layer, which may lose rich linguistic knowledge contained in the teacher network. Different from previous methods, we propose a many-to-many layer mapping method for BERT distillation, where each intermediate student layer can learn from any intermediate teacher layers adaptively. In addition, an Earth Mover's Disepstance is applied to learn the optimal many-to-many layer mapping solution.  

\begin{figure*}
    \centering
    \includegraphics[height=200pt]{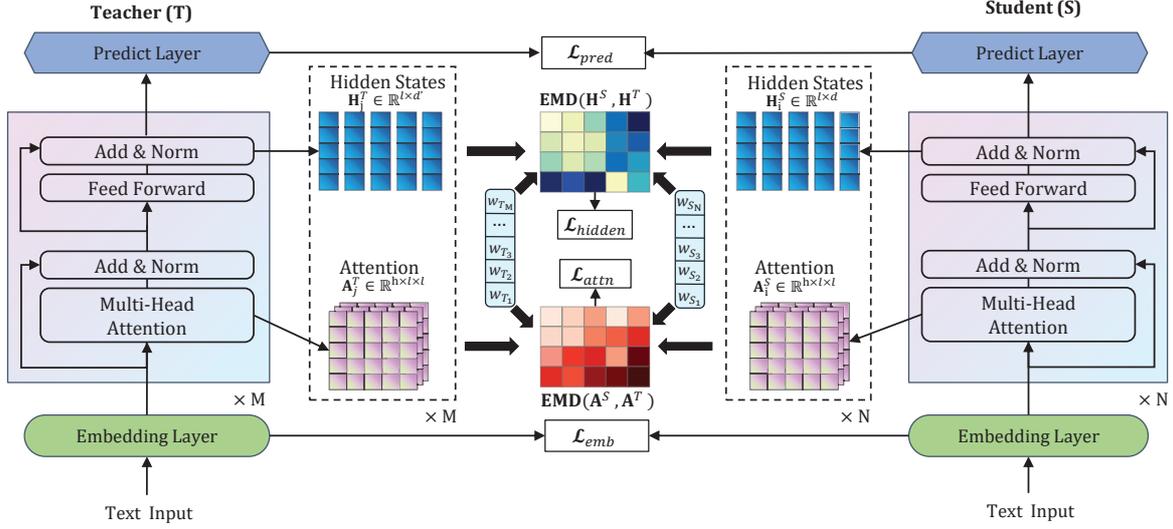}
    \caption{An overview of the proposed BERT-EMD method, which distills comprehensive knowledge from a large teacher ($T$) with $M$-layer Transformer to a small student ($S$) with $N$-layer Transformer. $w_{T_i}$ and $w_{S_j}$ are the weight of $i$-th teacher layer and $j$-th student layer used in EMD. Here, $l$ denotes the length of the input sequence. $h$ denotes the head number. $d$ and $d'$ are the hidden sizes of student and teacher Transformers, respectively.}
    \label{fig:weight}
\end{figure*}

\section{Methodology}
In this section, we propose a novel BERT compression method based on many-to-many layer mapping and Earth Mover's Distance (called BERT-EMD). In addition, we also propose a cost attention mechanism to 
learn the layer weights used in EMD automatically.

\subsection{Overview of BERT-EMD}
The main idea behind BERT-EMD is to transfer knowledge from a large teacher network $T$ (large BERT) to a small student network $S$ (BERT-EMD). Both the student and teacher networks are implemented with an embedding layer, several Transformer layers, and a prediction layer.
We assume that the teacher network has $M$ Transformer layers and the student network has $N$ Transformer layers.
Each Transformer layer contains an attention layer and a hidden layer. 

Similar to TinyBERT \cite{jiao2019tinybert}, our method also includes three primary distillation components: the embedding-layer distillation, the Transformer distillation, and the prediction-layer distillation. Concretely, both the embedding-layer distillation and the prediction-layer distillation employ the one-to-one layer mapping as in TinyBERT and BERT-PKD, where the two student layers are guided by the corresponding teacher layers, respectively. However, different from the previous works, we propose to exploit the many-to-many layer mapping for Transformer (intermediate layers) distillation (attention-based distillation and hidden states based distillation), where each student attention layer (resp. hidden layer) can learn from any teacher attention layers (resp. hidden layers). In this way, BERT-EMD can learn from different intermediate teacher layers adaptively for different NLP tasks, motivated by the intuition that different NLP tasks require different levels of linguistic knowledge contained in the attention and hidden layers of BERT. Next, we will describe the four distillation strategies of  BERT-EMD in detail.

\subsection{Embedding-layer Distillation}
Word embeddings are vital in NLP tasks and have been extensively studied in recent years. Better representations of words
have come at the cost of huge memory footprints. Compressing embedding matrices without sacrificing model performance is essential for real-world applications. To this end, we minimize the mean squared error (MSE) between the embedding layers of the teacher and  student networks:
\begin{equation}
\mathcal{L}_{\rm emb} = {\rm MSE}(\mathbf{E}^S \mathbf{W}_e, \mathbf{E}^T) 
\end{equation}
where the matrices $\mathbf{E}^S$ and $\mathbf{E}^T$ represent the embeddings of student and teacher networks, which have the same shape.  $\mathbf{W}_e$ is a projection parameter to be learned. 

\subsection{Prediction-layer Distillation}
The student network also learns from the probability logits provided by teacher network. We minimize the prediction-layer distillation function as:
\begin{equation}
\mathcal{L}_{\rm pred} = -\rm{softmax}(\mathbf{z}^T) \cdot \rm{log\_softmax}({\mathbf{z}^S / t})
\end{equation}
where $\mathbf{z}^T$ and $\mathbf{z}^S$ represent the probability logits predicted by the teacher and student, respectively. $t$ indicates a temperature value. 

\subsection{Transformer Distillation with Earth Mover's Distance}
Instead of imposing one-to-one layer mapping as in previous works \cite{sun2019patient,jiao2019tinybert}, our Transformer distillation approach allows many-to-many layer mapping and is capable of generalizing to various NLP tasks. 
The Earth Mover's Distance (EMD) is proposed to measure the dissimilarity (distance) between the teacher and student networks as the minimum cumulative cost of transforming knowledge from the teacher network to student network.  
The key insight is to view network layers as distributions, and the desired transformation should make the two distributions (teacher and student layers) close. 

\paragraph{Attention-based Distillation}
We use the attention-based distillation to transform the linguistic knowledge from the teacher network to the student network based on EMD. Formally, let $\mathbf{A}^T=\{(\mathbf{A}^T_1, w^\textbf{A}_{T_1}), \dots , (\mathbf{A}^T_{M}, w^\textbf{A}_{T_{M}})\}$  be the teacher attention layers and $\mathbf{A}^S = \{(\mathbf{A}^S_1, w^\textbf{A}_{S_1}), \dots , (\mathbf{A}^S_{N}, w^\textbf{A}_{S_{{N}}})\}$ be the student attention layers, where $M$ and $N$ represent the numbers of the attention layers in the teacher and student networks, respectively. 
Each $\textbf{A}^T_i$ (resp. $\textbf{A}^S_i$) represents the $i$-th teacher (resp. student) attention layer and $w^\textbf{A}_{T_i}$ (resp. $w^\textbf{A}_{S_i}$) indicates corresponding layer weight that is initialized as $\frac{1}{M}$ (resp. $\frac{1}{N}$). We also define a ``ground'' distance matrix  $\textbf{D}^\textbf{A}=[d^\textbf{A}_{ij}]$, where $d^\textbf{A}_{ij}$ represents the cost of transferring the attention knowledge 
from $\textbf{A}^T_i$ to $\textbf{A}^S_j$. Here, we use MSE to calculate the distance $d^\textbf{A}_{ij}$ as:
\begin{equation}
d^\textbf{A}_{ij}={\rm MSE}(\textbf{A}^S_i,\textbf{A}^T_j)    
\end{equation}

Then, we attempt to find a mapping flow $\mathbf{F}^{\mathbf{A}}=[f^{\mathbf{A}}_{ij}]$, with $f^{\mathbf{A}}_{ij}$ the mapping flow between $\mathbf{A}^T_i$ and $\mathbf{A}^S_j$, that minimizes the cumulative cost required to transform knowledge from the teacher attention layers $\mathbf{A}^T$ to the student attention layers $\mathbf{A}^S$:
\begin{equation}
{\rm WORK}(\mathbf{A}^T, \textbf{A}^S, \mathbf{F}^\textbf{A})  = \sum_{i=1}^{M} \sum_{j=1}^{N} {f^\mathbf{A}_{ij} d^\mathbf{A}_{ij}}
\end{equation}
subject to the following constraints:
\begin{align}
&f^\mathbf{A}_{ij} \ge 0 ~~~~~~ 1\leq i\leq M,~1 \leq j \leq N\\
&\sum_{j=1}^{N}{f^\mathbf{A}_{ij}} \leq w^\mathbf{A}_{T_i} ~~~~~~  1\leq i\leq M\\
&\sum_{i=1}^{M}{f^\mathbf{A}_{ij}} \leq w^\mathbf{A}_{S_j} ~~~~~~ 1 \leq j \leq N\\
&\sum_{i=1}^{M}\sum_{j=1}^{N}{f^\mathbf{A}_{ij}} = \min(\sum_{i=1}^{M}{w^{\mathbf{A}}_{T_i}}, \sum_{j=1}^{N}{w^{\mathbf{A}}_{S_j}})
\end{align}
where the first constraint forces the mapping flow to be positive. The second constraint limits the amount of attention information that can be sent by $\mathbf{A}^T$ to their weights. The third constraint limits the attention information that can be received by $\mathbf{A}^S$. The fourth constraint limits the amount of total flow.

The above optimization is a well-studied transportation problem \cite{hitchcock1941distribution}, which can be solved by previously developed methods \cite{rachev1985monge}. 
Once the optimal mapping flow $\mathbf{F}^{\mathbf{A}}$ is learned, we can define the Earth Mover's Distance as the work normalized by the total flow: 
\begin{equation}
{\rm EMD}(\textbf{A}^S, \textbf{A}^T) = \frac{\sum_{i=1}^{M}\sum_{j=1}^{N} {f^\textbf{A}_{ij} d^\textbf{A}_{ij}}}{\sum_{i=1}^{M}\sum_{j=1}^{N} {f^\textbf{A}_{ij}}} 
\end{equation}

Finally, the objective function for the attention-based distillation can be defined by the EMD between $\mathbf{A}^T$ and $\mathbf{A}^S$:
\begin{equation}
\mathcal{L}_{\rm attn} = {\rm EMD}(\mathbf{A}^S, \mathbf{A}^T)
\end{equation}

\paragraph{Hidden States-based Distillation}
Similar to attention-based distillation, we also learn the hidden layer mapping based on EMD. 
Formally, let $\mathbf{H}^T=\{(\mathbf{H}^T_1, w^\textbf{H}_{T_1}), \dots , (\mathbf{H}^T_{M}, w^\textbf{H}_{T_{M}})\}$  be the teacher hidden layers and $\mathbf{H}^S = \{(\mathbf{H}^S_1, w^\textbf{H}_{S_1}), \dots , (\mathbf{H}^S_{N}, w^\textbf{H}_{S_{{N}}})\}$ be the student hidden layers, where $M$ and $N$ represent the numbers of the hidden layers in the teacher and student networks, respectively. 
Each $\textbf{H}^T_i$ represents the $i$-th hidden layer and $w^\textbf{H}_{T_i}$ indicates corresponding layer weight that is initialized as $\frac{1}{M}$. We also define a ``ground'' distance matrix  $\textbf{D}^\textbf{H}=[d^\textbf{H}_{ij}]$, where $d^\textbf{H}_{ij}$ represents the cost of transferring the hidden states knowledge 
from $\textbf{H}^T_i$ to  $\textbf{H}^S_j$ and we use a learnable projection parameter as $\mathbf{W}_h$. MSE is applied to calculate the distance $d^\textbf{H}_{ij}$:
\begin{equation}
d^\textbf{H}_{ij}={\rm MSE}(\textbf{H}^S_i\mathbf{W}_h,\textbf{H}^T_j)    
\end{equation}

Then, a mapping flow $\mathbf{F}^{\mathbf{H}}=[f^{\mathbf{H}}_{ij}]$, with $f^{\mathbf{H}}_{ij}$ the mapping flow between $\mathbf{H}^T_i$ and $\mathbf{H}^S_j$, is learned by minimizing the cumulative cost required to transform knowledge from $\mathbf{H}^T$ to $\mathbf{H}^S$:
\begin{equation}
{\rm WORK}(\mathbf{H}^T, \textbf{H}^S, \mathbf{F}^\textbf{H})  = \sum_{i=1}^{M} \sum_{j=1}^{N} {f^\mathbf{H}_{ij} d^\mathbf{H}_{ij}}
\end{equation}
subject to the following constraints:
\begin{align}
&f^\mathbf{H}_{ij} \ge 0 ~~~~~~ 1\leq i\leq M,~1 \leq j \leq N\\
&\sum_{j=1}^{N}{f^\mathbf{H}_{ij}} \leq w^\mathbf{H}_{T_i} ~~~~~~  1\leq i\leq M\\
&\sum_{i=1}^{M}{f^\mathbf{H}_{ij}} \leq w^\mathbf{H}_{S_j} ~~~~~~ 1 \leq j \leq N\\
&\sum_{i=1}^{M}\sum_{j=1}^{N}{f^\mathbf{H}_{ij}} = \min(\sum_i^{M}{w^{\mathbf{H}}_{T_i}}, \sum_i^{N}{w^{\mathbf{H}}_{S_i}})
\end{align}

After solving the above optimization problem, we obtain the optimal mapping flow $\mathbf{F}^{\mathbf{H}}$. The earth mover's distance can be then defined as the work normalized by the total flow: 
\begin{equation}
{\rm EMD}(\textbf{H}^S, \textbf{H}^T) = \frac{\sum_{i=1}^{M}\sum_{j=1}^{N} {f^\textbf{H}_{ij} d^\textbf{H}_{ij}}}{\sum_{i=1}^{M}\sum_{j=1}^{N} {f^\textbf{H}_{ij}}} 
\end{equation}

Finally, the objective function for the hidden states-based distillation can be defined by the earth mover's distance between $\mathbf{H}^T$ and $\mathbf{H}^S$:
\begin{equation}
\mathcal{L}_{\rm hidden} = {\rm EMD}(\mathbf{H}^S, \mathbf{H}^T)
\end{equation}

\subsection{Weight Update with Cost Attention}
In the EMD defined in Section 3.4, each teacher layer (resp. student layer) is assigned an equal weight $w_T = \frac{1}{M}$ (resp. $w_S = \frac{1}{N}$).
Since different attention and hidden layers of BERT can learn different levels of linguistic knowledge, these layers should have different weights for various NLP tasks. 
Therefore, we propose a cost attention mechanism to assign weights for each attention and hidden layers automatically. 

The main idea behind the cost attention is to make the teacher and student Transformer networks be as close as possible. That is, we could reduce the overall cost of EMD by increasing the weights of the layers with low flow cost, while the weights of the layers with high flow cost should be decreased adaptively. 

We take the weight updating process of the teacher network as an example. The cost attention mechanism can be performed by three steps after learning the optimal solution (flow matrices $\mathbf{F}^{\mathbf{A}}$ and $\mathbf{F}^{\mathbf{H}}$ in EMD). First, we learn the transferring cost between each teacher and student layers (unit transferring cost). Formally, let $\bar{C^\textbf{A}_{T_i}}$ and $\bar{C^\textbf{H}_{T_i}}$ be the unit transferring cost of each attention and hidden layers respectively, which can be computed as:
\begin{gather}
\bar{C_{T_i}^\textbf{A}} = \frac{ \sum_{j=1}^{N}d^\textbf{A}_{ij}f^\textbf{A}_{ij} }{w_{T_i}}\label{eq:attention-cost}\\
\bar{C_{T_i}^\textbf{H}} = \frac{ \sum_{j=1}^{N}d^\textbf{H}_{ij}f^\textbf{H}_{ij} }{w_{T_i}}\label{eq:hidden-cost}
\end{gather}

Second, we update the weights ($w_{T_i}^\textbf{A}$ and $w_{T_i}^\textbf{H}$) of the teacher attention and hidden layers based on the learned unit transferring cost. Specifically, we compute the updated weights $\bar{w_{T_i}^\textbf{A}}$ and $\bar{w_{T_i}^\textbf{H}}$ as the inverse ratio of the transferring costs:
\begin{gather}
\bar{w_{T_i}^\textbf{A}} = \frac{\sum_{j=1} ^ {M}\bar{{C}_{j}^\textbf{A}}}{\bar{{C}_{T_i}^\textbf{A}}}\\
\bar{w_{T_i}^\textbf{H}} = \frac{\sum_{j=1} ^ {M}\bar{{C}_{T_j}^\textbf{H}}}{\bar{{C}_{T_i}^\textbf{H}}}
\end{gather}

Finally, we normalize the updated layer weights used in EMD via softmax, and introduce a temperature coefficient $\tau$ to smooth the results. In particular, we update weight $\bar{w_{T_i}}$ of the $i$-th Transformer layer used in EMD by averaging the corresponding weights of attention and hidden layers:
\begin{gather}
\bar{w_{T_i}} = \frac{1}{2}({\rm softmax}(\bar{w_{T_i}^\textbf{A}}/\tau) + {\rm softmax}(\bar{w_{T_i}^\textbf{H}}/\tau))
\end{gather}

It is noteworthy that the learned new weights are leveraged as the constrains to optimize the EMD problem in the next batch. Specifically, we initialize the $i$-th teacher attention and hidden layer weights ($w_{T_i}^{\mathbf{A}}$ and $w_{T_i}^{\mathbf{H}}$) in the $\eta$-th batch with the updated weight $\bar{w_{T_i}}$ learned in the $\eta-1$-th batch. In this way, we can further improve the performance of BERT-EMD and accelerate convergence time.

\subsection{Overall Learning Objective}
Finally, we combine the embedding-layer distillation, attention-based distillation, hidden states-based distillation, prediction-layer distillation objectives to form the overall knowledge distillation objective as follows:
\begin{equation}
\mathcal{L}_{\rm distill} = \beta (\mathcal{L}_{\rm emb} + \mathcal{L}_{\rm attn} + \mathcal{L}_{\rm hidden}) + \mathcal{L}_{\rm pred}
\end{equation}
where  $\beta$ is a factor that controls the weights of the three distillation objectives ($\mathcal{L}_{\rm emb}$, $\mathcal{L}_{\rm attn}$,  $\mathcal{L}_{\rm hidden}$).


\begin{table*}[]
\footnotesize
\centering
\setlength{\tabcolsep}{0.5mm}{
\begin{tabular}{l|cc|ccccccccc|l}
\hline
\thead{\textbf{Model}} & \textbf{Params}  & \textbf{Inference} & \textbf{MNLI-m} & \textbf{MNLI-mm} & \textbf{QQP} & \textbf{SST-2} & \textbf{CoLA} & \textbf{QNLI} & \textbf{MRPC} & \textbf{RTE} & \textbf{STS-b} & \textbf{AVE} \\ 
 & \textbf{Num}  & \textbf{Time} & (393k) & (393k) & (364k) & (67k) & (8.5k) & (108k) & (3.5k) & (2.5k) & (5.7k) & \\ \hline
$\rm {BERT_{BASE}}_{12}$-G & 110M & $\times$1 & 84.6 & 83.4 & 71.2 & 93.5 & 52.1 & 90.5 & 88.9 & 66.4 & 85.8 & 79.60 \\
$\rm {BERT_{BASE}}_{12}$-T & 110M & $\times$1 & 84.4 & 83.3 & 71.6 & 93.4 & 52.8 & 90.5 & 88.1 & 66.9 & 85.2 & 79.58 \\ 
\hline
$\rm {BERT_{SMALL}}_4$ & 14.5M & -  & 75.4 & 74.9 & 66.5 & 87.6 & 19.5 & 84.8 & 83.2 & 62.6 & 77.1 & 70.18 \\
$\rm DistillBERT_4$ & 52.2M & $\times$3.0 & 78.9 & 78.0   & 68.5 & \textbf{91.4} & \textbf{32.8} & 85.2 & 82.4 & 54.1 & 76.1 & 71.93 \\
$\rm BERT\text{-}PKD_4$  & 52.2M & $\times$3.0 & 79.9 & 79.3 & \textbf{70.2} & 89.4 & 24.8 & 85.1 & 82.6 & 62.3 & 79.8 & 72.60 \\
$\rm TinyBERT_4$ & 14.5M & $\times$9.4 & 81.2 & 80.3 & 68.9 & 90.0 & 25.3 & 86.2 & 85.4 & 63.9 & 80.4 & 73.51 \\ 
\textbf{BERT-EMD}$_4$ & 14.5M & $\times$9.4 & \textbf{82.1} & \textbf{80.6} & 69.3 & 91.0 & 25.6 & \textbf{87.2} & \textbf{87.6} & \textbf{66.2} & \textbf{82.3} & \textbf{74.66} \\ 
\hline
$\rm BERT\text{-}PKD_6$ & 66.0M & $\times $1.9 & 81.5 & 81.0 & 70.7 & 92.0 & 43.5 & 89.0 & 85.0 & 65.5 & 81.6 & 76.61 \\
$\rm BERT\text{-}of\text{-}Theseus_6$ & 66.0M & - & 82.4 & 82.1 & 71.6 & 92.2 & \textbf{47.8} & 89.6 & 87.6 & 66.2 & 84.1 & 78.18 \\
$\rm TinyBERT_6$ & 66.0M & $\times $1.9 & 84.4 & 83.1 & 71.3 & 92.6 & 46.1 & 89.8 & 88.0 & 69.7 & 83.9 & 78.77 \\ 
\textbf{BERT-EMD}$_6$ & 66.0M & $\times$ 1.9  & \textbf{84.7} & \textbf{83.5} & \textbf{72.0} & \textbf{93.3} & 47.5 & \textbf{90.7} & \textbf{89.8} & \textbf{71.7} & \textbf{86.8} & \textbf{80.00} \\ \hline
\end{tabular}
\caption{Experimental results on the GLUE test set. The subscript within each model name represents the number of Transformer layers. AVE represents the average score over all tasks. $\rm {BERT_{BASE}}_{12}$-G and $\rm {BERT_{BASE}}_{12}$-T indicate the results of the fine-tuned BERT-base from \cite{devlin2018bert} and in our implementation, respectively. } 
}
\label{table:result}
\end{table*}

\begin{table}[]
\small
\centering
\begin{tabular}{l|cccc}
\hline
   Method         & MNLI-m & QQP   & RTE  & STS-b \\ \hline
BERT-EMD$_4$ & 82.1   & 69.3  & 66.2 & 82.3 \\
~~~~w/o CA$_4$     & 81.6   & 69.0  & 65.1 & 81.6 \\
~~~~w/o EMD$_4$    & 80.7   & 67.7  &    64.1  & 80.7
\\\hline
BERT-EMD$_6$ & 84.7   & 72.0  & 71.7 & 86.8 \\
~~~~w/o CA$_6$      & 84.5   & 71.6  & 71.0   & 85.3\\ 
~~~~w/o EMD$_6$     & 84.2   & 71.2  & 70.4 & 84.7 \\ 

\hline
\end{tabular}
\caption{Ablation test results in terms of removing EMD (w/o EMD) and cost attention (w/o CA). }
\label{table:ablation}
    \vspace{-0.5cm}
\end{table}

\section{Experimental Setup}
\subsection{Experimental Data}
We evaluate our BERT-EMD model on the General Language Understanding Evaluation (GLUE) \cite{wang2018glue} benchmark, which is a collection of nine diverse sentence-level classification tasks. 
Concretely, GLUE consists of (i) Microsoft Research Paraphrase Matching (MRPC), Quora Question Pairs (QQP) and Semantic Textual Similarity Benchmark (STS-B) for paraphrase similarity matching; (ii) Stanford Sentiment Treebank (SST-2) for sentiment classification; (iii) Multi-Genre Natural Language Inference Matched (MNLI-m), Multi-Genre Natural Language Inference Mismatched (MNLI-mm), Question Natural Language Inference (QNLI) and Recognizing Textual Entailment (RTE) for natural language inference task; and (iv) the Corpus of Linguistic Acceptability (CoLA) for linguistic acceptability. 

\subsection{Evaluation Metrics}
Following previous works \cite{sun2019patient,jiao2019tinybert}, we use classification accuracy as the evaluation metric for SST-2, MNLI-m, MNLI-mm, QNLI, and RTE datasets. 
For a fair comparison with TinyBERT \cite{jiao2019tinybert}, the F1 metric is adopted for MRPC and QQP datasets, the Spearman correlation is adopted for STS-B, and the Matthew’s correlation is adopted for CoLA. The results reported for the test set of GLUE are in the same format as on the official leaderboard.

\subsection{Implementation Details}
Similar to TinyBERT, our BERT-EMD method also contains a general distillation and a task-specific distillation. In particular, we initialize our student model with the general distillation model provided by TinyBERT \footnote{https://github.com/TinyBERT/TinyBERT}.
The teacher model is implemented as a 12-layer BERT model ($\rm BERT_{{BASE}_{12}}$), which is fine-tuned for each task to perform knowledge distillation. 

We employ the grid search algorithm on the validation set to tune the hyper-parameters.
Since there are many hyper-parameter combinations, we first do the grid search on $\beta$ and the learning rate. Then, we fix the values of these two hyper-parameters and tune the values of the other hyper-parameters.
Specifically, the batch size is 32, the learning rate is tuned from $\{5e-5, 2e-5, 1e-5\}$, the parameter $t$ defined in Eq. (2) is tuned from $\{1,3,7,10\}$, the temperature coefficient $\tau$ is tuned from $\{1, 2, 5, 10\}$, and $\beta$ is tuned from $\{0.01, 0.001, 0.005\}$.

\subsection{Baseline Methods}
In this paper, we compare our BERT-EMD with several state-of-the-art BERT compression approaches, including the original 4/6-layer BERT models \cite{devlin2018bert},  DistilBERT \cite{tang2019distilling}, BERT-PKD \cite{sun2019patient}, TinyBERT \cite{jiao2019tinybert}, BERT-of-Theseus \cite{xu2020bert}. 
However, the original TinyBERT employs a data augmentation strategy in the training process, which is different from the other baseline models.
For a fair comparison, we re-implement the TinyBERT model by eliminating the data augmentation strategy.

It is noteworthy that we do not compare BERT-EMD with the recent MobileBERT \cite{sun2020mobilebert} and MiniLM \cite{wang2020minilm}, since MiniLM does not report the results on the GLUE test set and the MobileBERT model employs the Transformer block with different architectures. 

\section{Experimental Results}
\subsection{Main Results}
We summarize the experimental results on the GLUE test sets in Table 1.
The number below each task denotes the number of training instances. 
Following previous works \cite{sun2019patient,sun2019patient}, we also report the average values of these nine tasks (the ``AVE'' column). 
From the results, we can observe that BERT-EMD  substantially outperforms state-of-the-art baseline methods by a noticeable margin on most tasks. 

Among all the 4-layer BERT approaches, our $\rm BERT\text{-}EMD_4$ method achieves the best results on almost all the tasks except SST-2 and CoLA. 
First, $\rm BERT\text{-}EMD_4$ achieves significantly better results than $\rm BERT_{{SMALL}_4}$ on all the GLUE tasks with a large improvement of 4.48\% on average. 
Second, $\rm BERT\text{-}EMD_4$ also outperforms $\rm DistilBERT_4$ and BERT-PKD$_4$ by a substantial margin, even with only 30\% parameters and inference time. 
Furthermore, $\rm BERT\text{-}EMD_4$ exceeds the TinyBERT model (the best competitor) by 2.3\% accuracy on RTE, 2.2\% F1 on MRPC, and 1.9\% Spearman correlation on STS-B. 
This verifies the effectiveness of our BERT-EMD model in improving the performance of small BERT-based methods on various language understanding tasks. 

We can observe similar trends in the 6-layer BERT models.
Table 1 shows that the proposed $\rm BERT\text{-}EMD_6$ method can effectively compress $\rm BERT_{{BASE}_{12}}$ into a 6-layer BERT model without performance sacrifice.
Specifically, $\rm BERT\text{-}EMD_6$ performs better than the 12-layer BERT
$\rm BERT_{{BASE}_{12}}$ model on 7 out of 9 tasks, with only about 50\% parameters and inference time of the original $\rm BERT_{{BASE}_{12}}$ model. For example, BERT-EMD achieves a noticeable improvement of 5.3\% accuracy on RTE and 1\% Spearman correlation on STS-B, over the $\rm BERT_{{BASE}_{12}}$ model.

\begin{figure*}
    \centering
    \includegraphics[height=210pt]{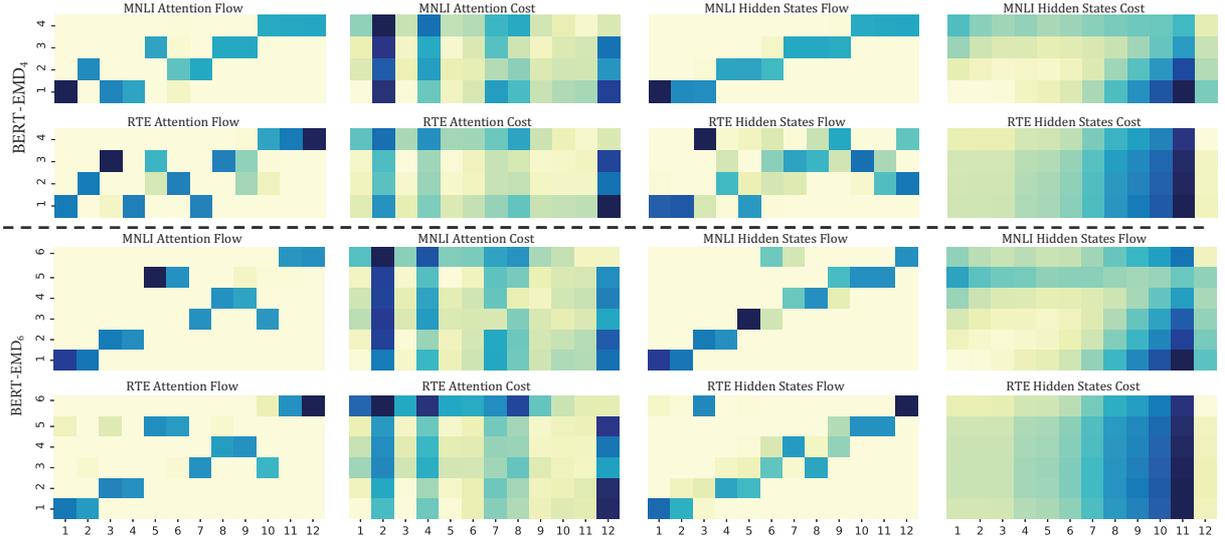}
    \caption{The visualization of flow matrices ($\mathbf{F}$) and distance matrices ($\mathbf{D}$) in developing BERT-EMD$_4$ (above) and  BERT-EMD$_6$ (below) for two examples from MNLI and RTE tasks, respectively. The abscissa represents the Transformer layers of $\rm BERT_{{BASE}_{12}}$, and the ordinate represents the Transformer layers of BERT-EMD$_4$/BERT-EMD$_6$. The color depth represents the values (weights) of the layers.}
    \label{fig:heatmap}
    \vspace{-0.5cm}
\end{figure*}

\subsection{Ablation Study}
To verify the effectiveness of  EMD and the cost attention mechanism, we perform ablation test of BERT-EMD on two large datasets (MNLI and QQP) and two small datasets (MRPC and RTE) in terms of removing EMD (denoted as w/o EMD) and cost attention (w/o CA), respectively. In particular, for the method of removing EMD, we retain the many-to-many layer mapping by simply replacing the EMD with the mean squared error when measuring the distance between the teacher and student layers.

The ablation test results are summarized in Table \ref{table:ablation}. Generally, both EMD and cost attention contribute noticeable improvement to our method. 
The performances decrease sharply, especially on the STS-B task, when removing the EMD module. This is within our expectation since the EMD module formulates the distance between the teacher and student networks as an optimal transport problem, which helps to learn an optimal many-to-many layer mapping. 
The cost attention also contributes to the effectiveness of BERT-EMD. This verifies that the cost attention can further improve the many-to-many layer mapping by learning the importance of each teacher layer in guiding the student network.
It is noteworthy that when removing the EMD module in the many-to-many lay mapping process, our w/o EMD$_4$ performs slightly worse than TinyBERT$_4$ on the MNLI and QQP tasks. This is because we cannot automatically control the information flow during the many-to-many layer mapping without using EMD, which further verifies the effectiveness of EMD in the many-to-many layer mapping process.

\subsection{Visualization of Compression Process}
To better understand the  many-to-many layer mapping process, we illustrate the flow matrices $\mathbf{F}$ and cost (distance) matrices $\mathbf{D}$ in developing BERT-EMD$_4$ (above) and  BERT-EMD$_6$ (below) for two examples from MNLI and RTE tasks, respectively. 
In Figure \ref{fig:heatmap}, we report the averaged values of the flow and cost matrices of the entire epoch that achieves the best performance on the validation set with heat maps.

From the results in Figure \ref{fig:heatmap}, we have several key observations. First, different tasks could emphasize different teacher layers in compressing the Transformer. The diagonal positions of the matrices are almost always important for the MNLI task, which exhibits similar trends with TinyBERT with the one-to-one ``Skip'' layer mapping strategy. However, for the RTE task, each student Transformer layer can learn from any teacher Transformer layers. The previous one-to-one layer mapping methods cannot take full advantage of the teacher network. This argument can be verified by the quantitative results in Table 1, where our BERT-EMD has a much larger improvement on RTE than on MNLI over TinyBERT. 
Second, comparing BERT-EMD$_4$ and BERT-EMD$_6$, we can observe that BERT-EMD$_4$ usually needs to learn more comprehensive information from skipped teacher Transformer layers, resulting in more divergent many-to-many layer mappings. 


\section{Conclusion}
In this paper, we propose a novel BERT compression method based on many-to-many layer mapping by Earth Mover's Distance (EMD). To our knowledge, BERT-EMD is the first work that allows each intermediate student layer to learn from any intermediate teacher layers adaptively. 
In addition, a cost attention mechanism is designed to further improve the model's performance and accelerate convergence time by learning the layer weights used in EMD automatically.  
Extensive experiments on GLUE tasks show that BERT-EMD can achieve competitive performances with the large BERT-Base model while significantly reducing the model size and inference time.

\section*{Acknowledgement}
Min Yang was partially supported by National Natural Science Foundation of China (No. 61906185), the Natural Science Foundation of Guangdong Province of China (No. 2019A1515011705, 2018A030313943), the Youth Innovation Promotion Association of CAS.

\end{document}